# Representing Knowledge Base into Database for WAP and Web-based Expert System

Istiadi
Departement Of Electrical Engineering
Widyagama University of Malang
Malang, Indonesia
e-mail: istiadi@widyagama.ac.id

Emma Budi Sulistiarini
Departement Of Industrial Engineering
Widyagama University of Malang
Malang, Indonesia
e-mail: emma@widyagama.ac.id

*Abstract*— Expert System is developed as consulting service for users spread or public requires affordable access. The Internet has become a medium for such services, but presence of mobile devices make the access becomes more widespread by utilizing mobile web and WAP (Wireless Application Protocol). Applying expert systems applications over the web and WAP requires a knowledge base representation that can be accessed simultaneously. This paper proposes single database to accommodate the knowledge representation with decision tree mapping approach. Because of the database exist, consulting application through both web and WAP can access it to provide expert system services options for more affordable for public.

## I. INTRODUCTION

Expert system as a part of artificial intelligence has potential application on the various fields [1]. Concept of an expert system is to map the human expertise into a computer system to find the solution in a problem domain. Users can get advantage of expert system consulting services on their problem [2]. Problem domain could include target for specific users or the public which concerned on these problems. When the target of an expert system is to services widely users, the accessibility services need special attention.

Presence of the Internet as computers network that ability to distribute information widely. The application of expert systems over the internet using a web application has strived as in [3, 4, 5]. But it is not sufficient, because maybe not everyone can reach internet connection through direct access. An alternative that would expand access of expert system services can be done through mobile devices (cell phones, smartphones, etc.). Today's mobile devices have ability accessing the web created with responsive design or access through Wireless Application Protocol (WAP), so that access can be done.

Web and WAP applications utilize the different base language, namely HTML and WML, so the application of expert systems that using them require different markup language platform. Manage two applications for the same content would be less efficient if done separately. Main content of the expert system is the knowledge base. So, it requires an approach of how to represent and accommodate knowledge representation through medium that can be accessed by both platforms.

This paper proposes a model of expert systems with knowledge representation expressed using the decision tree approach that mapped into database. So that consulting application services on web and WAP platforms can access it as a front-end applications. Existence of the database, the application developer can be implemented using a web-based to manage the knowledge representation of expert system remotely.

## II. RELATED WORK

Expert System is a tool for non-expert users to obtain support finding solution in a certain domain of problem that mapping the human expert knowledge into a computer program [2]. Initially computer is working standlone, but now growth in the form of a network of local scale until Internet established. Web-based application over the Internet has provided alternative expert system services development. Among the advantages of web-based expert system is from the aspect of its information distrution but it become more complex [6]. The Development of web-based expert system requires attention for user interaction. The system should be able to generate web pages dynamically for the user towards the inference process [3].

Development of expert systems using web requires medium to express knowledge representation. In [3], an expert system is developed using Java (JSP) that utilizes JESS as a module for the formalization of knowledge. In (4), a web-based expert system have developed using database to store data related symptoms, diseases, and treatment was given but not yet accommodate the base rules. In [5], a web-based expert system is utilized for assessment services suppliers with tree-diagram approach that are similar to the decision tree.

The presence of mobile devices that accommodate the data communication system has become the options in accessing such as web. A data communications standard for mobile devices such as WAP has been developed [7]. This provides chance for expansion the services of expert systems applications. Utilization expert systems through WAP has been proposed in [8, 9], but still using static approach that maps the decision tree as WAP pages that linked each other. It complicated when updating knowledge representation.

Based on both web based and WAP based services, they need the similar content of knowledge representation. Therefore, a medium need to provide to be used shareable. In this case, the knowledge representation can be expressed in the form of decisien tree. While both based on support accessing the database using a web programming language, the efforts are needed to map the decision tree into database and then how it interprets by both types of services.

### III. PROPOSED MODEL

The proposed model try to accommodate the differences both platforms web-based consulting services and WAP-based consulting services through database to store the knowledge representations. With the decision tree approach, the knowledge representation need mapped to the form of database that allow accessible by both platforms. Further design architecture consisting of development application based on web and consulting application based on web and WAP that used it.

#### A. Mapping Decision Tree to Database

Concept of a decision tree is concist of nodes that are connected as shown in Fig. 1. A root node as initial facts to begin identification process. The decision node is a node that has a branch that cover the conditional are fulfilled, The leaf nodes as end position is generally the decisions or conclusions that lead to a particular solution.

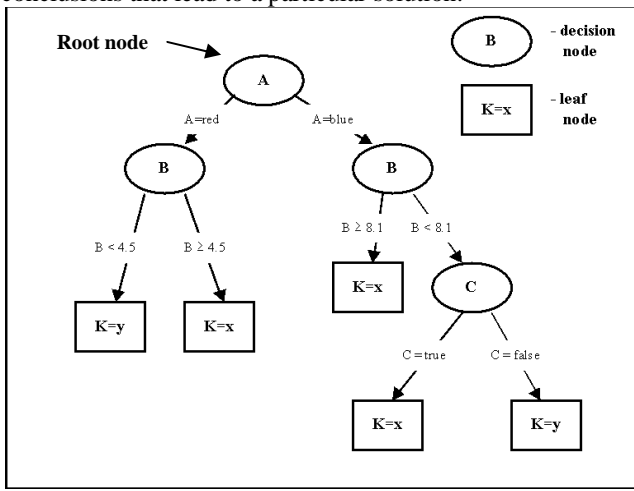

Figure 1. Example of decision tree [10]

Based on the description of the decision tree concept there needs to:
- Defining the scope or domain problems (cases) that concern.
- Defining the symptoms that may arise in the identification process.
- Defining the conclusion as result of the diagnosis of symptoms that appear.
- Defining the rules base date that describes the relationship between the nodes as a representation symptom identification or representation conclusion.

Consider with the identification of these needs, database model was developed in the form of relasionship models (Fig. 2.)

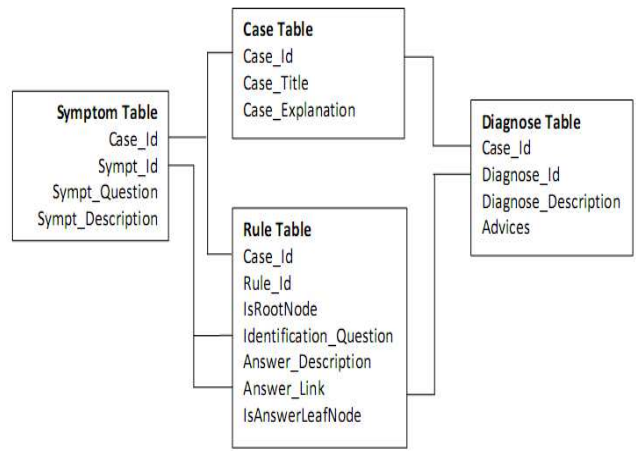

Figure 2. Design of databases to accommodate knowledge representation

Figure 2 shows several tables to capture the knowledge representation of expert system. Case table accommodates the data surrounding the case definition of the problem domain. Symptoms tables store the definition of symptoms that includes questions to identify symptoms and it description. Diagnosis table is a medium to store data of conclusions definition or diagnose result and suggestions or advices were given. Rules table to accommodate the data of tree structure representation that consisting of initial identification stages data (to determine the root node or decision node), identification of symptoms data (to refer the question), the data description of the answer as a statement of the relation between nodes, the data refer to answer options to identification of symptoms or conclusions (diagnosis), and the end identification data (IsAnswerLeafNode) that constitute the answer as indication of further identification (symptoms) or finish at a conclusion (diagnose).

#### B. Design of System Architecture

System architecture was designed to cover users to access the expert system consulting services in web forms over the internet and the users who access through mobile devices such as smartphones using the mobile internet (mobile web) and mobile phone may access using WAP applications. Entry of expert system knowledge data are necessary on web-based applications development in order to be accessed remotely. This is because the web application is sufficient to give the appearance of features on-line management. Fig. 3 shows the architecture of WAP and web based expert system.

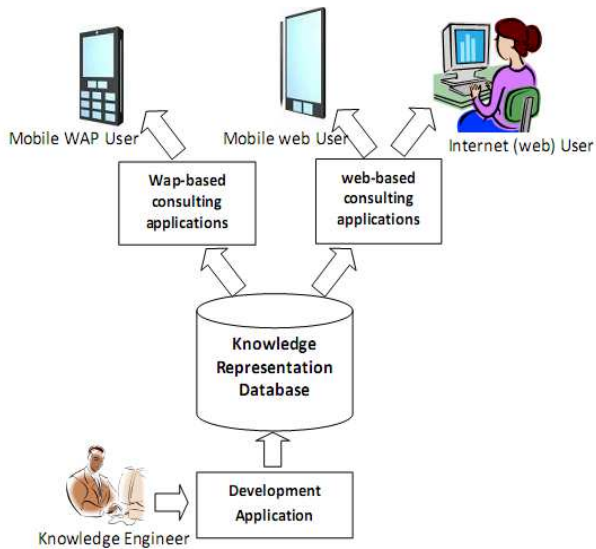

Figure 3. Proposed WAP and Web based expert system architecture

The main part of system is the database that holds the mapping of knowledge representation. From side of application developer, knowledge engineer roles to entry or editing of data stored in the knowledge database as defining of case (problem domain), description of symptoms, description of diagnosis, and formulate the rules. It is like role of web site administrator that manages by Conten Management System (CMS). While the conten of knowledge representation has been accommodated completely, then the user can be access it using the consulting application. Although these have different palform, but they have the similar interpretation to work using these databases

. The interpretations expressed in the consulting application algorithm as follows :
1. Based on the case or problem domain that selected, do the query of records on the rules tables that indicates the root node
2. Generate identification (question) page by showing :
   a. Identification questions by querying the symptoms table where symptom_id similar to the symptoms identification field on the rule tables that selected.
   b. Choice of answers would refer (hyperlink) to the next branch based on each record selected from the rule table that contains element of answers types (which can be either a decision node or leaf node) and the answer option (which can be either symptom_id or diagnose_id)
3. Based on the answer selected by the user :
   a. If type of answer is decision node, then do the query record on the rules table that the field of identification_question corresponding to the selected symtom_id. Next do step 2.
   b. If type of answer is leaf node, then do the query record on the diagnose table based on the selected diagnose_id field and generate a page that displays the conclusions (diagnose results) and suggestions were given.

IV. CASE EXAMPLE : DENGUE FEVER CONSULTING

Case examples used in this paper is a consulting service Dengue disease. This disease is one disease that is still a concern because it often leads to death. According to the WHO [11] between 2000 and 2007 occurred more than 900,000 cases in more than 60 countries. Even in Indonesia, according to the health ministry occurred more than 150,000 cases by the number who died more than 1,300 people in 2009 [12]. The occurrence of deaths generally occur because of late handling of patients.

Emergency action need to be supported in addition to prevention efforts. Provision of information interactively (consultative) to act the incidence of the disease is necessary for the public such as online utilize the consulting application of expert systems. In [13], an expert system has developed for dengue fever but for local application as an early detection tool for physicians, while the public may be necessary for early detection of the symptoms are easily identified.

WHO have agreed classification of cases of dengue fever [11], ie without warning sign dengue, dengue with warning signs, and severe dengue. Each of these classifications has symptoms that indicate diagnosis and what suggestions actions need to be done. Based on these references, the application is directed to an expert system consulting dengue fever and illness levels to identify necessary actions. Identification was based on the facts that readily perceived or recognized in patients such as fever, rash, nausea, joint pain, bleeding, weak pulse, and loss of consciousness.
.

V. RESULT AND DISCUSION

WAP and web-based expert system developed aims to cover the needs from both side of development (expert or knowledge engineer) and consulting (users). Applications were built using the PHP programming language and MySQL database. Moreover web-based application used responsive design that can be adapt the screen size of user devices [14].

*A. Web based Development Application*

Development application only for specific users ( expert or knowledge engineer ) to entry or editing the knowledge representation data into database. Therefore, these services need to provide authentication interface. Once inside within application development, users could define the problem domain (case) that it scope the expert system. Furthermore definition of the symptoms could be done as a statement of identification in the form of questions.

Another definition was diagnose description as a result of inference and a statement of recommendation or advice provided as alternative offered solutions to users pertaining to the conclusions. After the symptoms that may arise and the possible diagnosis results have been defined, then developers could define the rules relating to occurance of symptoms directed to the results of conclusion (diagnosis).

Defining the rules provided through the service of entry / editor as shown in Fig. 4.

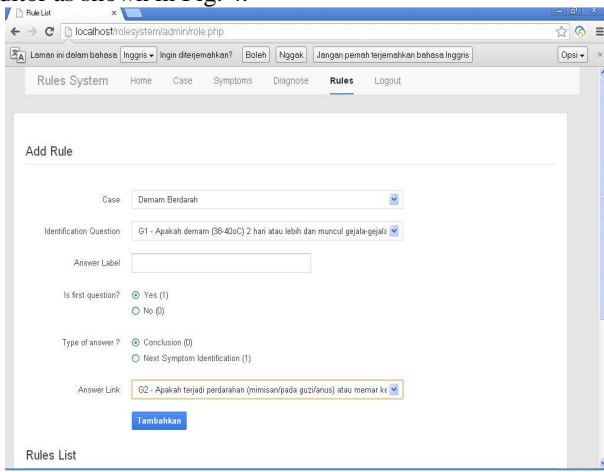

Figure 4. Display of Entry or Editor Form of defining rules

Fig. 4 show many items to facilitate creation of rule-based by decision tree approach. The identification question should be chosen from combobox that containt list of questions and next the answer label must be defined too. Based on decision tree approach, initial identificataion question is declared as root node that can be chosen from options (Yes or No) of Is First Question item. Then the answer links must be selected that posible consist of the next identification question or the conclusion (result of diagnose). Furthermore, developer must set the type of answer is decision node or leaf node.

After whole contents of the knowledge base has been declared, then the application of consulting services can be utilized by users. In this case users can access the application through the internet (web) or through the mobile devices (eg mobile phones and smartphones) which is connected.

### B. Web based Consulting Application

Web-based consulting applications to provide services user access to the internet (web). Applications created with responsive design to adapt to mobile web users with smaller screen. Fig. 5 is an example of page identification consultation view.

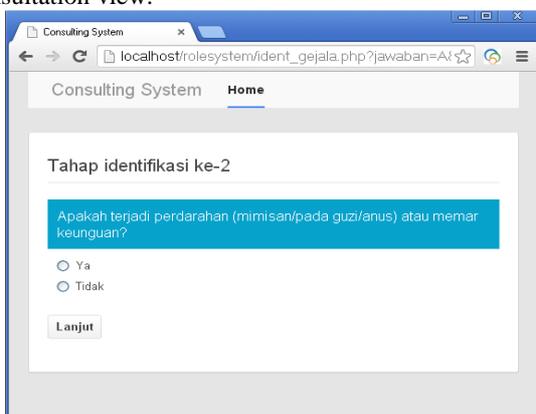

Figure 5. Examples of web-based identification (question) page

In the consultation process, a question is given as a representation of the symptoms identification. Users can choose the answer which is available by selecting one item on the answer options. To continue the process, the user is asked to press (click) the Continue button. The identification process will take step by step to obtain a decision or conclusion. Fig. 6 show an example of conclusion page.

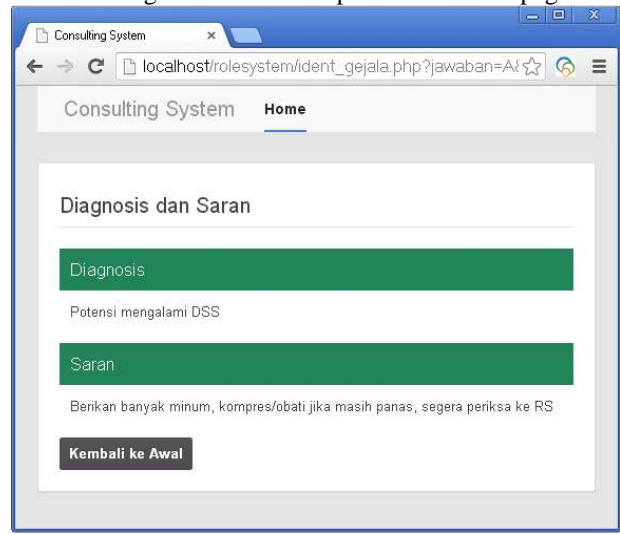

Figure 6. Example of web-based conclusion (diagnosis and advice) page

Fig. 6 exemplified a conclusion page which is contains the results of diagnosis and advice. Diagnosis is the conclusion of the stages prior identification. Advice or recommendation given as an alternative solution to the problem based on the conclusion (diagnosis) which is obtained.

### C. WAP based Consulting Application

WAP-based expert system application is developed to the consulting application services which are intended for users who will access using mobile phones. This is to accommodate users of mobile phones that do not support access via the web but using WAP. Fig. 7 show some examples of the WAP-based consulting application display that was developed.

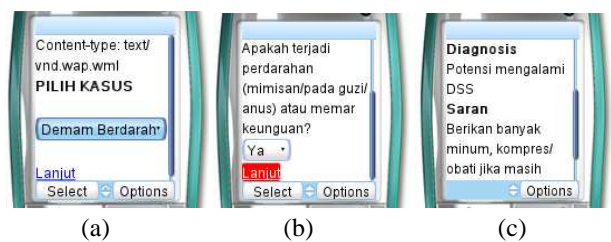

(a)         (b)         (c)

Figure 7. Examples display of WAP based consulting application (a) first page (b) identification (question) page (c) conclusion page

Just as web-based application, a WAP-based consulting application services will be run a symptom identification process step by step. The application will display the home page (Fig. 7a) when the service starts accessed. After user presses the Continue link, the process continued with the identification process through the generation of the identification page (Fig. 7b), which contains questions related symptoms that may arise. Users are asked to answer the questions according to the responses and continue the process by pressing the Continue link. After identification phase has leads to a conclusion then conclusion page will be raised. Conclusion page (Fig. 7c) contains the results of diagnosis and advice who required by users.

## VI. Conslusion

Model of expert system knowledge representation in the form of a decision tree enabling accommodated in a database that consists of tables to store data of case definition, symptoms definitions, diagnose definition and rules definition. The database that holds the knowledge representation, then development application can be made for entry or editing knowledge base and rule base for knowledge engineer, and consulting applications services can be developed for user to access trought web which responsive design (for mobile web) and trought WAP that have reference to the database.

### Acknowledgment

This paper is part of a research grant from the Directorate General of Higher Education, Ministry of Indonesian Education.